# Sharing standardized image-derived data in computational pathology using DICOM

Daniela P. Schacherer (1)*, Christopher P. Bridge (2, 3)*, David Clunie (4), Igor Octaviano (5), André Homeyer (1), Markus D. Herrmann (3, 6), Olivier Gevaert (7), Tabita Ghete (8), Markus Metzler (8), Henning Hoefener (1), Tahsin Kurc (9), Curtis Lisle (10), Kenneth Philbrick (11), Joel Saltz (9), Yuanning Zheng (7), Andrey Fedorov (3, 12)

*contributed equally
(1) Fraunhofer Institute for Digital Medicine MEVIS, Bremen, Germany
(2) Athinoula A. Martinos Center for Biomedical Imaging, Massachusetts General Hospital, Boston, USA
(3) Harvard Medical School, Boston, USA
(4) PixelMed Publishing, Bangor, USA
(5) Radical Imaging LLC, USA
(6) Massachusetts General Hospital, Boston, USA
(7) Stanford University, Stanford, USA
(8) University Hospital Erlangen, Erlangen, Germany
(9) Stony Brook University, Stony Brook, USA
(10) National Institute of Allergy and Infectious Diseases, Bethesda, USA
(11) Google Research, Mountain View, USA
(12) Brigham and Women's Hospital, Boston, USA)

## Abstract

Development and evaluation of computational pathology methods require access to large and diverse datasets. Over the past decade, various initiatives invested significantly into collecting, centralizing, and sharing pathology imaging data. In contrast, sharing of image-derived data such as region-of-interest delineations or segmentation masks is less well developed. In this work, we describe our approach to encoding and sharing image-derived pathology data in a standardized manner within the National Cancer Institute (NCI) Imaging Data Commons (IDC), a platform that hosts and provides public access to de-identified radiology and pathology data. The IDC relies on the Digital Imaging and Communications in Medicine (DICOM) standard for data harmonization, yet the adoption of DICOM for pathology image-derived content has remained largely unexplored until now. Here, we present five representative datasets harmonized by conversion from their original representations into DICOM and shared publicly in the IDC. We demonstrate the benefits of this harmonization, describe contributions to critical open-source tooling, and discuss technical considerations relevant to broader adoption of DICOM for pathology image-derived data.

## Introduction

Computational methods in pathology enable scientists and clinicians to analyze large whole slide images (WSIs) at scale, with considerable promise for accelerating biomedical research, improving diagnosis, and developing personalized treatment regimens. Efficient and reproducible development and evaluation of such methods requires access to large and diverse datasets. Several initiatives have therefore started to collect and share pathology image data across institutions, including The Cancer Genome Atlas Program (TCGA)[1], The Clinical Proteomic Tumor Analysis Consortium (CPTAC)[2], and the Human Tumor Atlas Network (HTAN)[3,4].

The existence of various data collection initiatives creates a need for standardized platforms that simplify sharing and use of pathology imaging data according to the Findability, Accessibility, Interoperability, and Reusability (FAIR) principles[5]. One such platform is the National Cancer Institute (NCI) Imaging Data Commons (IDC)[6], which as of Spring 2026 hosts approximately 100 TB of radiology and pathology cancer image data. The IDC relies on the Digital Imaging and Communications in Medicine (DICOM) standard to define the data model and representation of stored images and metadata, as well as the DICOMweb interface[7] for accessing them.

Beyond sharing the images, it is important to share information derived from them. Image-derived data can appear in various forms including slide-level labels (e.g., diagnoses), delineated regions of interest (ROIs) (e.g., outlining specific tissue types or cells), dense pixel-level classifications (e.g., segmentation masks), and quantifications or categorizations thereof. The data may be curated manually by experts (expert annotations), automatically generated by computational pathology tools, or a combination of both. Access to expert annotations is essential for developing and evaluating image analysis algorithms that perform, for example, automatic detection or delineation of certain tissue types[8,9], cell nuclei segmentation and classification[10,11], or tumor microenvironment characterization[12–14]. Self- and weakly-supervised learning methods have somewhat reduced the need for expert annotations, yet they remain indispensable for fine-tuning and validation and can significantly lower the total amount of data required for training[9]. Sharing analysis results produced by computational tools facilitates their reuse, benchmarking, and refinement, allowing others to build on existing insights rather than starting from scratch.

Whereas image-derived data are currently shared by individual research groups in a variety of general-purpose formats, we describe here our approach to standardize sharing of such data within the IDC using DICOM. This choice offers several advantages. Having a single standard that governs both file format and metadata conventions across modalities and domains supports the implementation of the FAIR principles. It facilitates interoperability and provides the foundation for unified programming interfaces, with the practical consequence of being able to use open-source DICOM tools for tasks such as archival, search, efficient access, and visualization, independent of dataset-specific variations. This, in turn, simplifies maintenance, improves sustainability, and reduces risks associated with relying on infrastructure maintained by a single entity. Ultimately, all of this can enhance the reproducibility of data-driven research[15].

Although encoding of image-derived information in DICOM is being increasingly explored in radiology[16–20], and recent extensions allow such encoding for pathology[19,21–23], to our knowledge, no public pathology datasets of this kind exist (outside of IDC). Our work addresses this gap by releasing exemplar collections of image-derived pathology data in DICOM and developing practical infrastructure as resources for those using, creating, or converting such data in the future.

The specific contributions of this manuscript are as follows. First, we describe our approach to encoding image-derived pathology data using DICOM and to making these data accessible through the IDC, illustrated by five specific example datasets we have converted from their original representations. Second, we highlight the capabilities that are enabled by the standardized DICOM representation, including improved searchability, visualization, and programmatic access through a uniform interface. In that context, we provide open-source notebooks and a dashboard that demonstrate how different tools can operate with the resulting DICOM collections. Third, we describe developments on two critical open-source tools, namely a Python library for creating and parsing image-derived data objects and web viewer. Finally, we discuss technical considerations and trade-offs involved in using DICOM to encode image-derived data in pathology.

# DICOM for encoding data derived from whole slide images

This section provides background information on the capabilities that DICOM offers for encoding and representing images and image-derived data in digital pathology.

DICOM defines application-specific representations of medical imaging data and their metadata in Information Object Definitions (IODs). Here, we briefly summarize the basic concepts of using DICOM for digital pathology, and refer to Herrmann et al.[21] for further details. The Visible Light Whole Slide Microscopy Image IOD supports high-resolution WSIs with specimen and patient metadata. To manage the large file sizes inherent to WSIs, an image is stored as a tiled multi-resolution pyramid, with each level of the pyramid being encoded as a distinct DICOM instance (file) of the same DICOM series. Within each instance, the large pixel array, referred to as *total pixel matrix*, is divided into smaller tiles that are stored as individually compressed frames. This organization allows, for example, viewing software to render lower-magnification views without accessing the full-resolution data. In addition, two options for the arrangement of tiles, known as Dimension Organization Types, are defined: the *tiled-full* representation places tiles in an implicit, pre-determined order, whereas the *tiled-sparse* representation explicitly defines the location of every tile. The sparse representation permits arbitrary ordering and the omission of empty tiles, but requires a non-trivial expansion of the metadata to record the tile positions[21,24].

Beyond the storage of WSIs, the DICOM standard defines multiple IODs that collectively allow for a variety of common image-derived data in pathology to be encoded. In this paper, we consider four IODs for image-derived data: Segmentation (SEG), Parametric Map (PM), Structured Reporting (SR), and Microscopy Bulk Simple Annotations (MBSA, often referred to as ANN). These objects share with all other DICOM objects their low-level encoding and contextual metadata attributes, including identifiers and descriptors of the patient and the imaging study, dates and times, acquisition equipment details, and, where applicable, clinical context. Each IOD then uses a further set of attributes to encode the image-derived information itself, and descriptive metadata concerning its generation and meaning (see Figure 1). For encoding metadata, DICOM makes extensive use of standardized terminologies to convey concepts unambiguously. A cell nucleus, for example, would be expressed not as the plain English string “nucleus”, but rather as coded concept combining the code value “84640000”, its meaning “Nucleus”, and the respective terminology, here the Systematized Nomenclature of Medicine – Clinical Terms (SNOMED CT or SCT)[25].

Segmentation objects store segmentations of other DICOM images as raster (pixel-based) masks containing one or more distinct regions, referred to as *segments*. Each segment is described by coded concepts specifying what it represents, for example, a tissue region, cell type, or subcellular structure. Like WSIs, segmentation masks can be tiled using either the tiled-full or tiled-sparse representation, and can, but need not, be encoded as a multi-resolution pyramid[26]. Although it is a common special case, the segmentation masks do not need to be represented at the same resolution(s) as the pyramid levels of the source image(s). The SEG IOD defines three different types of segmentations, distinguished by the value of the `SegmentationType` attribute (`BINARY`, `LABELMAP`, and `FRACTIONAL`). *Binary* and *labelmap*[27] segmentations are used for cases where each pixel either belongs to a segment or does not. Binary segmentations use a separate binary raster mask for each segment. A labelmap instead uses a single mask, in which each pixel value identifies the segment to which that pixel belongs, optionally including a background segment. Binary segmentations allow overlapping segments, i.e., a pixel can belong to more than one segment. They are, however, less space efficient than labelmap segmentations when many segments are present. *Fractional* segmentations are used when a pixel has partial membership of a segment, with either a probabilistic or occupancy interpretation, represented using a continuous value for each pixel and a separate raster mask per segment.

Parametric Map objects store pixel arrays of image-derived measurements along with coded concepts for the measured quantity and its unit[28]. One key advantage of PMs over other DICOM objects, which only store pixels using integer data types, is the fact that they can also store pixel values in 32-bit or 64-bit floating point format, allowing for a higher level of precision. Like SEGs, they can be stored in a tiled format using either the tiled-full or tiled-sparse representation, with or without additional multi-resolution pyramid layers.

Two further object types allow storage of image-derived information in non-raster formats. The first, Structured Reporting[29], defines a broad class of objects that can store various types of content related to images within a tree-structured hierarchy. This can include quantitative measurements, qualitative evaluations as free text or coded concepts, as well as spatial regions defined by vector graphic data such as points, circles, or polygons. SR templates constrain the virtually unlimited flexibility of the SR encoding for specific applications. For our purposes here, we consider only SRs following Template 1500 “Measurement Report” (TID1500)[30]. This general-purpose template stores image-derived findings (quantitative measurements and qualitative evaluations) organized into *measurement groups* that each relate to an image or image region, and is widely used in radiology applications[16,31]. However, because each spatial region is accompanied by a considerable number of attributes describing its meaning and provenance, storing large numbers of similar regions in an SR quickly becomes impractical, given that, for example, automatically generated cell boundaries in a pathology image can easily number in the tens of thousands.

The second IOD was designed specifically to address such cases: the Microscopy Bulk Simple Annotations object[32]. Each MBSA object contains one or more *annotation groups*, each of which encodes a set of similar structures, with metadata the same across the set. Within an annotation group, graphic data are stored efficiently in large arrays, optionally accompanied by arrays of corresponding per-region numerical measurements. The MBSA IOD is also distinguished by being specific to pathology applications, which are unique in the potentially extremely large number of individual annotations per image (hence “bulk”), whereas PMs, SEGs, and SRs are equally applicable in – and in fact were originally designed for – radiology and other areas of medical imaging.

**DICOM Segmentation (SEG)**

.dcm
SOPInstanceUID:
1.2.826.0.1.3680043.8.498...
--
Segment
Segment Number:
1
Segmented Property Category:
(91720002, SCT, "Body substance")
Segmented Property Type:
(181769001, SCT, "Connective tissue")
Segment Algorithm Type:
AUTOMATIC
Segmentation Algorithm Identification Sequence:
Algorithm Name:
Tissue-Type-Seg
Columns
Rows
Total Pixel Matrix
Image Coordinate System
[px]
**Raster data**
(stored in Image Pixel module)

**DICOM Microscopy Bulk Simple Annotations (MBSA)**

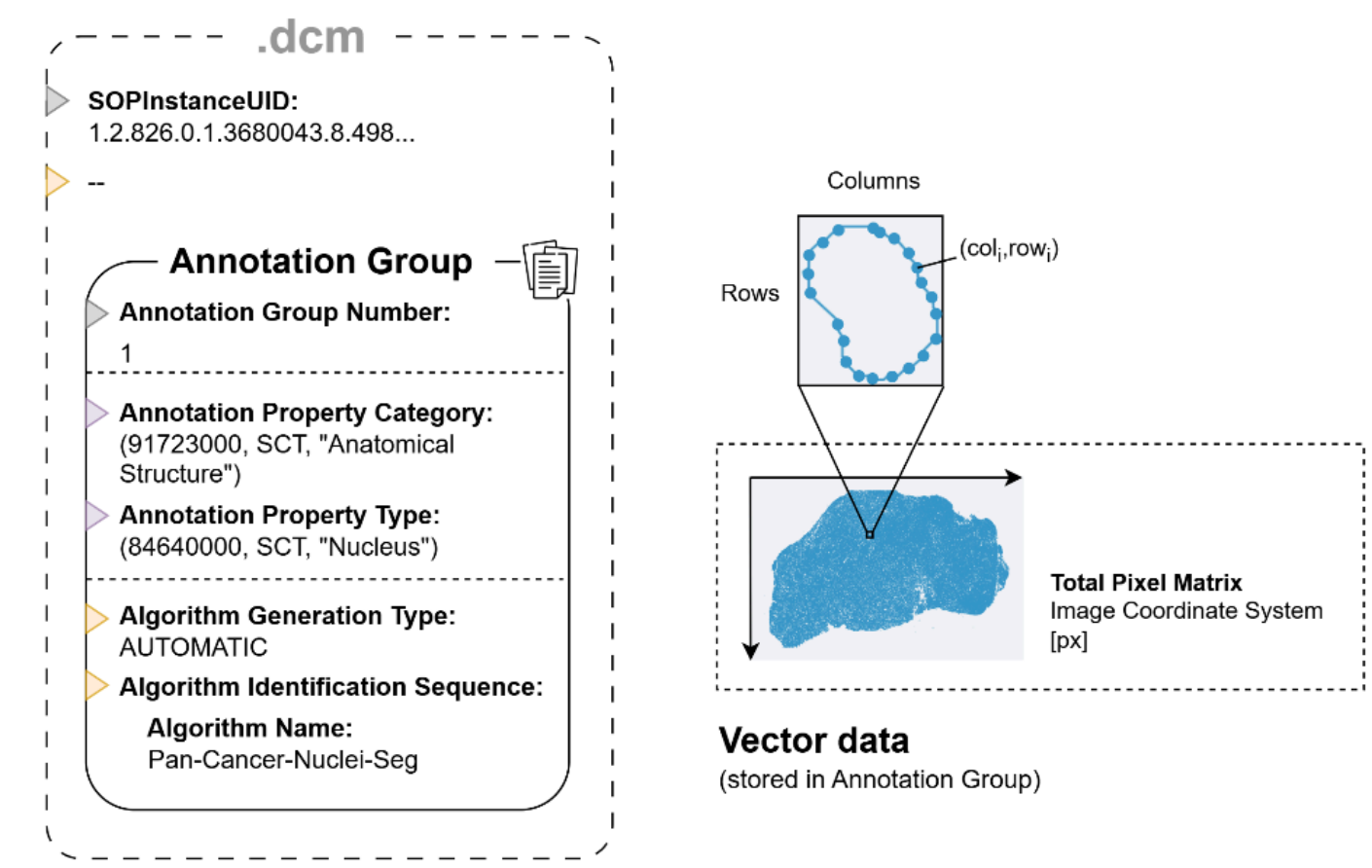

**DICOM Parametric Map (PM)**

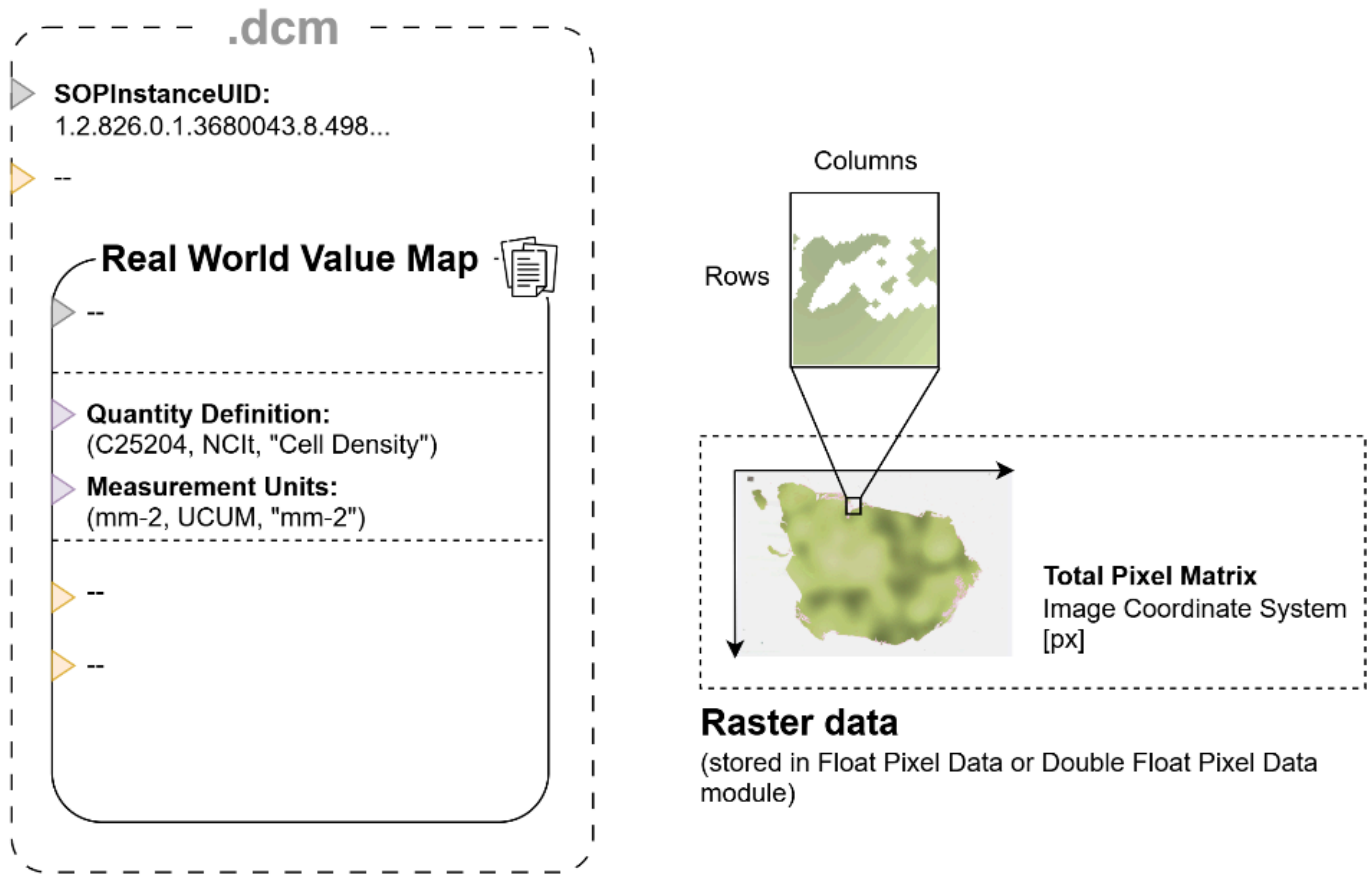

**DICOM Structured Reporting (SR)**

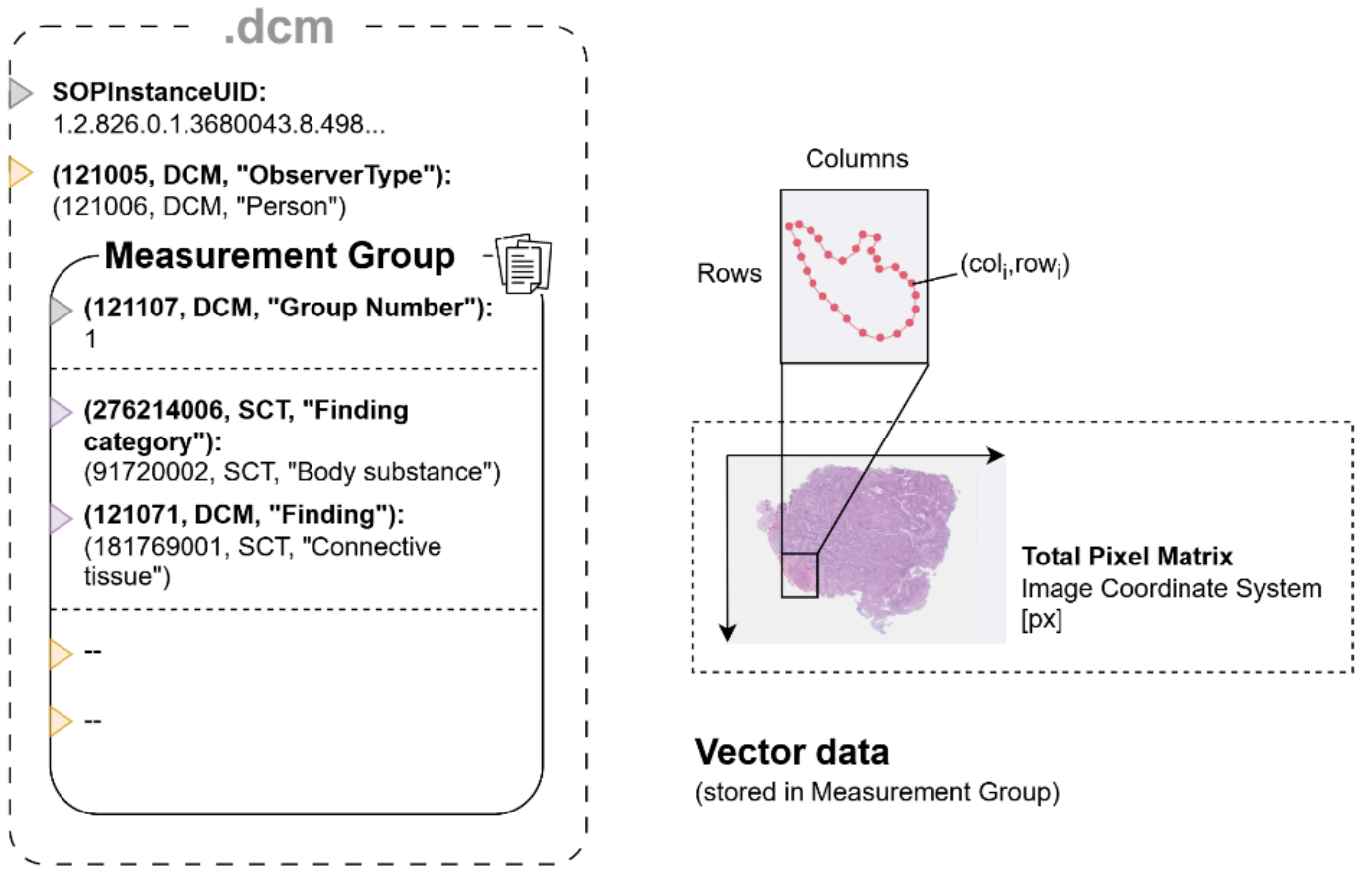

**Figure 1: Schematic representation of four DICOM Information Object Definitions (IODs) suitable for encoding image-derived data from WSIs.** Shown are Segmentation (SEG), Microscopy Bulk Simple Annotations (MBSA), Parametric Map (PM), and Structured Reporting (SR), covering both raster-based (SEG, PM) and vector-based (MBSA, SR) representations. For each IOD, the dashed box on the left represents one instance of that IOD, and lists important attributes related to data identification (grey), meaning (violet) and algorithm provenance (yellow). The inner box separates instance-level attributes from those defined at the level of the individual encoded unit (segment, annotation group, real world value map, measurement group) of which an instance can contain multiple (stack icon). Exemplary values are given for each attribute, including coded concepts (code value, coding scheme, code meaning). To the right of each box, the spatial data contained in that instance are shown in their position on the total pixel matrix, with a magnified view above illustrating the encoded content in more detail.

# Methods

This section describes the work that was done to harmonize five independent sets of pathology image-derived data into DICOM and to make them publicly accessible via the IDC. The datasets were prioritized based on our assessment of their value to the cancer research community. We only considered data that the creators were willing to share under a permissive CC-BY license.

In the following, we first provide an overview of the technical implementation that is common to all dataset conversions. Then, for each dataset, we detail the background and provenance, content, size, original format, chosen DICOM representation, and dataset-specific conversion details. Table 1 summarizes this information. Finally, we describe the validation procedures applied to all resulting DICOM instances, and the infrastructural work for data hosting, access, and visualization. The criteria for choosing the most suitable DICOM representation for each dataset are presented in the Results section, because the selection process evolved iteratively throughout this work.

## DICOM harmonization of image-derived data

Dataset conversions were implemented using custom Python code building on, among others, the open-source `pydicom`[33], `numpy`[34], and `highdicom`[19] packages. The `highdicom` library was developed by the IDC team specifically to read and create DICOM objects encoding image-derived data, including SEG, PM, SR and MBSA IODs. `highdicom` handles many low-level tasks including input validation, generation of DICOM unique identifiers, linking to source images via their identifiers, and the inheritance of context attributes, such as patient-, study-, and specimen-level information, from the source images. We curated the remaining attributes specific to the image-derived data, including provenance information such as algorithm details or the producer, and passed them to the `highdicom` API. For semantic information on spatial regions or accompanying measurements we used codes from established terminologies, including SNOMED CT, the NCI Thesaurus (NCIt)[35], and DICOM's own controlled terminology (DCM)[36].
The Python code for the dataset conversions is publicly available[37,38]. Throughout this work, we made several important enhancements to `highdicom` (see Enhancements to open-source tooling in the Results section) in order to simplify similar processes for others in the future.

### Cell nuclei contours (Pan-Cancer-Nuclei-Seg-DICOM)

This dataset contains automatically-generated contours outlining cell nuclei in over 6000 hematoxylin and eosin (H&E) stained WSIs from 14 different cancer types within TCGA (BLCA, BRCA, CESC, COAD, GBM, LUAD, LUSC, PAAD, PRAD, READ, SKCM, STAD, UCEC, UVM) collections. The data were generated by a segmentation model trained on synthetically created images and segmentation masks[39,40], and were previously made publicly available in a CSV text file format through The Cancer Imaging Archive (TCIA)[41,42]. There are multiple CSV files per slide, each covering a specific slide region and containing the polygon coordinates and area in pixels for all detected cell nuclei. We obtained information on how to interpret the CSV files, including coordinate system and file structure, from an accompanying README file and the TCIA webpage.

During conversion, we stored the polygon coordinates in one MBSA instance per WSI. Since all contours outline the same type of object, we stored them in the same annotation group. We used SNOMED CT terminology to describe the objects' semantics. Specifically, we assigned 'Anatomical Structure' (SCT code 91723000) as `AnnotationPropertyCategory` and 'Nucleus' (SCT code 84640000) as `AnnotationPropertyType`. The pixel-based nuclei areas were converted into physical areas in square microns using the slide-specific pixel spacing, and stored in the `MeasurementsSequence`. To facilitate consumption by raster image processing tools, we additionally rasterized the contours into segmentation masks, and stored these as a tiled multi-resolution pyramid of binary SEGs matching the pyramid levels of the source image pyramid. We used the same SNOMED CT codes as in the MBSA instances for `SegmentedPropertyCategory` and `SegmentedPropertyType` in the SEG objects.

### Rhabdomyosarcoma tissue typing (RMS-Mutation-Prediction-Expert-Annotations)

This is a dataset of tissue type delineations for 96 H&E slides of rhabdomyosarcoma (RMS), a rare and aggressive soft tissue cancer with two major molecular subtypes, embryonal rhabdomyosarcoma (ERMS) and alveolar rhabdomyosarcoma (ARMS). Tissue type data from two sources were available: 1) coarse region-level contours manually created by a human expert and used for training a deep learning segmentation model, and 2) finer-grained model predictions generated by that segmentation model[43]. The same four tissue types (stroma, necrosis, ERMS, and ARMS) are delineated in both cases. The manual annotations were provided as contours in XML format from the proprietary Aperio ImageScope annotation tool[44]. By contrast, the high resolution model predictions were provided as probabilistic raster masks in the NumPy array serialization format. In both cases, information on how to interpret the data was obtained through personal communication with the dataset authors (Milewski *et al.*, manuscript in preparation).

We stored the expert-generated contours in one SR per annotated slide, using the TID1500 Measurement Report template. For each region, we captured semantic information in `FindingCategory` and `Finding`, with 'Body substance' (SCT code 91720002) or 'Morphologic abnormality' (SCT code 49755003) for the former and 'Connective tissue' (SCT code 181769001), 'Necrosis' (SCT code 6574001), 'Alveolar rhabdomyosarcoma' (SCT code 63449009), or 'Embryonal rhabdomyosarcoma' (SCT code 14269005) for the latter. The model predictions were stored in a tiled multi-resolution pyramid of fractional SEGs, reusing these codes as values for `SegmentedPropertyCategory` and `SegmentedPropertyType`.

### Tumor-infiltrating lymphocytes detection (TCGA-SBU-TIL-Maps)

This dataset contains patch-level classifications indicating the presence or absence of tumor-infiltrating lymphocytes (TILs) in H&E WSIs from 23 collections within TCGA (ACC, BLCA, BRCA, CESC, COAD, ESCA, HNSC, KIRC, LIHC, LUAD, LUSC, MESO, OV, PAAD, PRAD, READ, SARC, SKCM, STAD, TGCT, THYM, UCEC, UVM). The data were generated by deep learning models and made publicly available as part of two distinct data releases accompanying two publications. The first publication[13] used a custom CNN model to generate classifications for a subset of 13 TCGA collections, making the results available in TCIA[45] in raster Portable Network Graphic (PNG) format files. This first set of data uses two classes: 'TILs present' and 'TILs absent' for regions of tissue that do not contain TILs (as distinct from the background class). The second publication[46] used an improved version of the model able to run on more tumor types, and therefore classifications are available for 23 collections within TCGA. These data use a single 'TILs present' class with no explicit 'TILs absent' class, and were made publicly available in a custom text file (TXT) format containing the coordinates of patches analyzed and their associated probabilistic model predictions and binarized labels[47]. For both data releases, information on how to interpret patch coordinates and predicted values was obtained from the publications and the TCIA webpage.

We converted both sets to SEGs. For the first set, this involved converting the raster array from its original PNG to the SEG format, whereas for the second set an initial step required creating rasterized masks from the provided TXT files. We created binary segmentations for both sets and additionally fractional segmentations for the second set. Because the segmentation masks consist of one pixel for an entire patch in the source image, they are inherently low resolution. Therefore, the use of a multi-resolution pyramid was not required. We used the SNOMED CT code 'Morphologically abnormal structure' (SCT code 49755003) for `SegmentedPropertyCategory` and either 'Tumor infiltration by lymphocytes present' (SCT code 399721002) or 'Tumor infiltration by lymphocytes absent' (SCT code 396396002) for `SegmentedPropertyType`.

### Glioblastoma tissue characterization (TCGA-GBM360)

This dataset contains patch-level maps of cancer aggressiveness scores calculated by a neural network for glioblastoma slides within the TCGA-GBM[48] collection. The maps were generated as part of a research study that used deep learning-derived histopathologic features to predict prognosis with the open-source GBM360 tool created by the authors for this purpose[49]. The raw data were supplied to us as arrays of patch coordinates and accompanying aggressiveness scores (with values between 0 and 1) in Hierarchical Data Format (HDF5) files. Information describing the intended interpretation of patch coordinates and aggressiveness scores was gathered from the publication and through personal communication.

The patch coordinates were used to form raster arrays of aggressiveness scores using NumPy, which were then encoded in one PM per slide. Due to the low resolution of the maps, a multi-resolution pyramid was not required. As no standardized code existed for the concept of aggressiveness, we introduced the private code 'Aggressiveness score' (private 99PMP code 314001), and used it with the DICOM attribute `QuantityDefinitionSequence` to capture the semantics of the values. We used the Unified Code for Units of Measure (UCUM)[50] to specify the values' unit in the `MeasurementUnitsCodeSequence`, in this case a dimensionless value normalized to the interval [0,1], expressed as UCUM annotation {0:1}.

### Pediatric leukemia bone marrow region and cell annotations (BoneMarrowWSI-PediatricLeukemia)

This dataset, described in detail in the corresponding preprint[11], contains manually generated ROIs and cell annotations for 246 bone marrow aspirate smear WSIs of pediatric patients which were diagnosed with either acute lymphoid leukemia (ALL), acute myeloid leukemia (AML), or chronic myeloid leukemia (CML). The rectangular ROIs outline parts of the evaluable monolayer area that are suitable for diagnosis because cells are evenly dispersed and minimally overlapping. Within the ROIs, experts marked hematological structures, predominantly immune cells, with bounding boxes. For a subset of these structures, they additionally assigned free-text labels from a set of 49 distinct terms, including specific cell types such as lymphocyte, or neutrophilic metamyelocyte, but also terms like spicule, mitosis, or damaged cell. Labels were assigned asynchronously per structure, with one annotator contributing a single label per iteration (annotation session) until a consensus label was established or disagreement was deemed stable. All data were available to us in CSV files that contained the coordinates of the bounding boxes, numeric identifiers and in case of labeled hematological structures all labels assigned as part of the consensus labeling approach. How to interpret bounding box coordinates and label assignments was partially documented in the provided CSVs and supplemented through personal communication with the dataset authors.

For each annotated slide, we created one MBSA series containing the ROIs and one series containing annotations of the hematological structures. Where these structures were labeled, we created one series per annotation session. For the ROIs, we used 'Spatial and relational concepts' (SCT code [309825002]) as `AnnotationPropertyCategory` and 'Selected region' (DCM code [111099]) as `AnnotationPropertyType`. Unlabeled hematological structures were assigned 'Anatomical structure' (SCT code [91723000]) as `AnnotationPropertyCategory` and 'Structure of hematological system' (SCT code [414387006]) as `AnnotationPropertyType`. For the labeled structures, we selected suitable codes from SNOMED CT, NCIt and Logical Observation Identifiers Names and Codes (LOINC)[51] in collaboration with the clinical experts (see Table S1). We used the `MeasurementsSequence` attribute to store the numeric identifier for each ROI, and, for each hematological structure, both its own numeric identifier and a reference to the ROI containing it. `ClinicalTrialSeriesID` was used to encode the annotation session. More details on this specific dataset conversion can be found in the technical documentation attached to this collection's Zenodo record[52].

**Table 1: Key characteristics of the five pathology image-derived source datasets.** Abbreviations: IOD = information object definition; MBSA = Microscopy Bulk Simple Annotations; PM = Parametric Map; ROI = region of interest; SEG = Segmentation; SR = Structured Report; TCIA = The Cancer Imaging Archive; TID = template identifier; TIL = tumor-infiltrating lymphocytes.

| | **Cancer type** | **Content** | **Generation type** | **Source format** | **Metadata source** | **DICOM target IOD(s)** |
|---|---|---|---|---|---|---|
| Cell nuclei contours | Pan-cancer (14 types) | Cell nuclei contours | automatic | CSV | Dataset README, TCIA webpage | MBSA, additionally SEG |
| Rhabdomyosarcoma tissue typing | Rhabdomyosarcoma | Tissue type region contours | manual and automatic | XML (Aperio ImageScope), NumPy | Personal communication | SR (TID1500), SEG |
| Tumor-infiltrating lymphocytes detection | Pan-cancer (23 types) | Patch-level TIL presence (binary and continuous) | automatic | PNG, TXT | Paper, TCIA webpage | SEG (binary + fractional) |
| Glioblastoma tissue characterization | Glioblastoma | Patch-level aggressiveness scores (continuous) | automatic | HDF5 | Paper, personal communication | PM |
| Pediatric leukemia bone marrow region and cell annotations | Leukemia | ROI and cell bounding boxes partially with labels over time | manual | CSV | CSV, personal communication | MBSA |

## Validation

This work focuses on the harmonization of the source datasets into a uniform DICOM representation – not on the generation of image-derived content *de novo*. Therefore, the intent of the validation step is to ensure that the resulting datasets comply with the DICOM standard, and that the user-facing tools operate as expected with the generated data. For scientific validation of the data, see the references cited for each source dataset.

We relied on the validators from `dicom3tools`[53] to confirm DICOM standard compliance. Specifically, we used `dciodvfy` to verify conformance of the created files to their respective IODs and `dcentvfy` to ensure consistency of series-, study- and patient-level attributes across instances of the same DICOM series. For SRs we used the `DicomSRValidator` from the PixelMed Java DICOM Toolkit[54]. Any non-compliance issues resulted in an iterative review of the DICOM standard's text, improvements to the conversion tools, and where newer DICOM capabilities were not yet fully covered, to the validators themselves. Following successful validation, the files were ingested into a Google Healthcare DICOM Store and visualized with the Slim[55] viewer, as described below. Only when these two steps were completed satisfactorily, we proceeded with staging the resulting image-derived DICOM files for ingestion into the IDC (see below). After release, we spot-checked the datasets in the production environment.

## Infrastructure and access

The final, validated DICOM files were ingested into the IDC. As a result, the converted data became accessible from cloud buckets via S3 and other APIs on two independent platforms (Amazon Web Services and Google Cloud), and from a Google Healthcare DICOM Store via DICOMweb. Fine-grained selection of individual instances is possible with the BigQuery Structured Query Language (SQL) interface, which offers access to all of the DICOM metadata automatically extracted from the DICOM files, or with the Python library `idc-index`[56], which exposes a curated subset of these attributes. Notably, the IDC maintains all its data in a single DICOM store, and relies on a single BigQuery table for all metadata across radiology and pathology images and image-derived data. In addition, the entirety of data is available through the IDC Portal web interface.

## Visualization

IDC users can visualize the converted image-derived data collections using the Slim microscopy viewer. Slim retrieves both image and non-image objects over the standard DICOMweb interface and renders them client-side in the browser. In the course of this work, we significantly improved Slim's capabilities for handling image-derived DICOM objects (see Enhancements to open-source tooling in the Results section).

# Results

First, this section introduces the image-derived data collections in the IDC that resulted from harmonizing the source datasets into DICOM. It then demonstrates practical improvements to their usability achieved through this harmonization process. The remainder of the section describes enhancements to open-source tools, and discusses technical aspects and considerations to be made when using DICOM to encode image-derived pathology data, as well as limitations encountered.

## DICOM-converted data collections

The harmonized image-derived data were ingested into the IDC and made available alongside the source images, as summarized in Table 2. Figure 2 displays representative examples of source images and their derived content. The IDC organizes data submissions that include images into *original collections*. These may also contain derived data prepared by the submitting entity. Subsequently generated image-derived data contributed at a later time, often by entities other than the original submitters, are typically placed in *analysis result collections*.

**Table 2: Overview of the DICOM-converted image-derived data collections in IDC v24.** *The complete list of collections is given in the respective Methods subsections. [1] We identified nine duplicated SEG series and ten duplicated MBSA series. [2] Because of corrupted tiles, one slide and its annotations could not be converted. Abbreviations: MBSA = Microscopy Bulk Simple Annotations; PM = Parametric Map; SEG = Segmentation; SR = Structured Report; TID = template identifier.

| | **IDC (original or analysis result) collection containing image-derived content** | | | | |
|---|---|---|---|---|---|
| | **Pan-Cancer-Nuclei-Seg-DIC** | **RMS-Mutation-Predictio** | **TCGA-SBU-TIL-Maps** | **TCGA-GBM360** | **BoneMarrowWSI-PediatricLeu** |

| | OM | n-Expert-An notations | | | kemia |
|---|---|---|---|---|---|
| IDC original collection(s) containing the analyzed images | 14 TCGA collections* | RMS-Mutation-Prediction | 23 TCGA collections* | TCGA-GBM | BoneMarrowWSI-PediatricLeukemia |
| Annotated patients | 5185 | 96 | 7600 | 336 | 245 |
| Annotated slides | 6065 | 97 | 8367 | 691 | 245[2] |
| DICOM SEG series | 6074[1] | 97 | 21030 | – | – |
| DICOM MBSA series | 6075[1] | – | – | – | 1027 |
| DICOM SR TID1500 series | – | 96 | – | – | – |
| DICOM PM series | – | – | – | 691 | – |
| Total size (GB) | 6970.57 | 6.98 | 2.04 | 0.36 | 0.02 |
| First added in IDC release (year) | v19 (2024) | v18 (2024) | v23 (2025) | v23 (2025) | v24 (2026) |

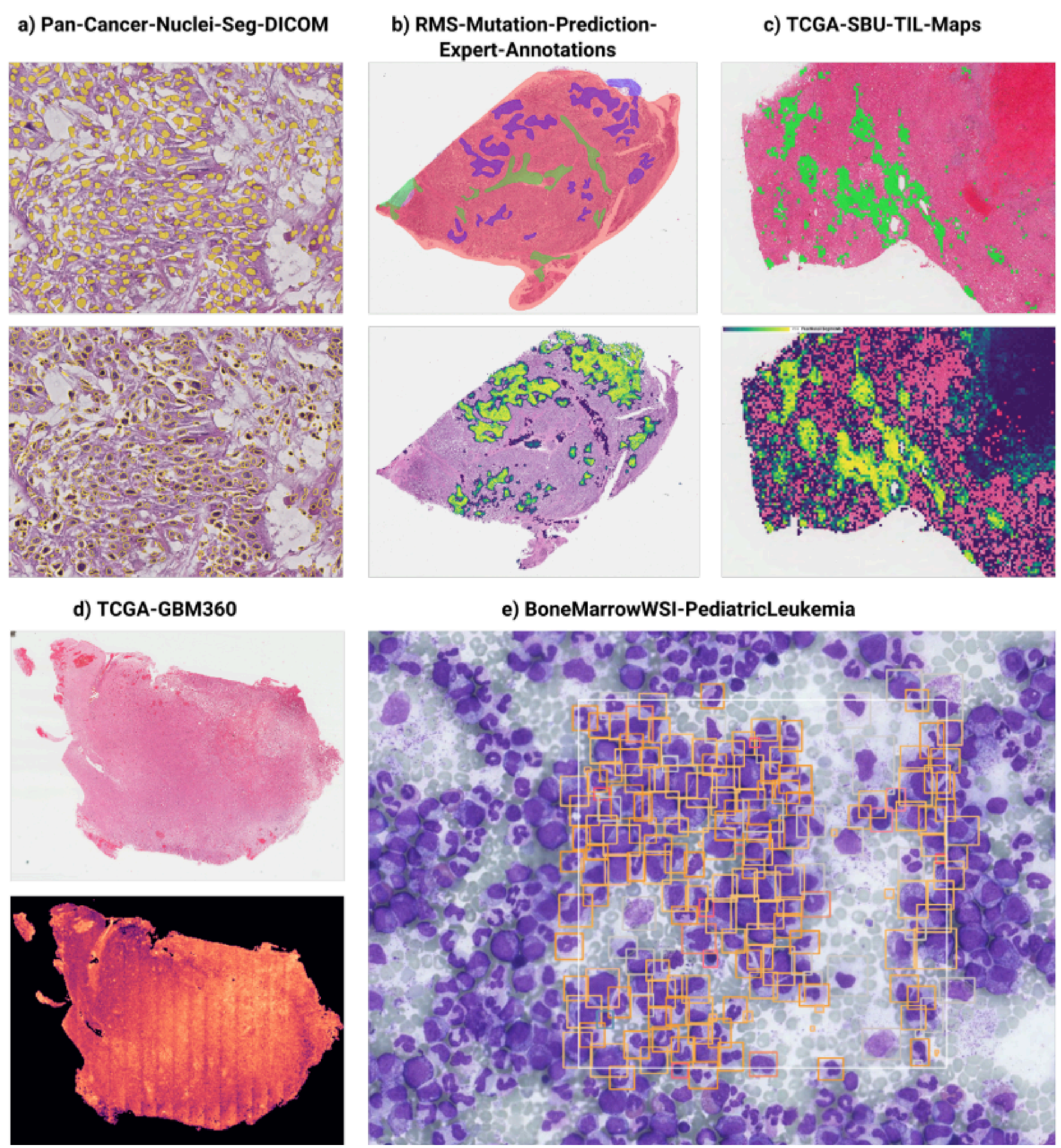

**Figure 2**: **Examples of images and image-derived data from the converted collections, visualized using the Slim viewer.** a) Region of an image from the TCGA-BLCA collection overlaid with cell nuclei contours (yellow) from the Pan-Cancer-Nuclei-Seg-DICOM collection, encoded as DICOM SEG (above) and DICOM MBSA (below). b) Image from the RMS-Mutation-Prediction collection overlaid with manual region contours from the RMS-Mutation-Prediction-Expert-Annotations collection, encoded as DICOM SR (above). The contours show alveolar rhabdomyosarcoma (red), stroma (green), and necrosis (blue). The same image is overlaid with a machine learning model's prediction of the tissue type encoded as fractional SEG (below). c) Image from the TCGA-LIHC collection overlaid with model-generated predictions of TIL-positive patches from the TCGA-SBU-TIL_Maps collection, encoded as binary SEG (above) and as fractional SEG (below). d) Region of an image from the TCGA-GBM collection without image-derived data (above) and with an overlaid tumor aggressiveness map from the TCGA-GBM360 collection stored as DICOM PM (below). e) Region of an image from the BoneMarrowWSI-PediatricLeukemia collection showing an expert-drawn ROI and cell bounding boxes stored in a DICOM MBSA. Links to the images shown are provided in the Supplementary Information. Abbreviations: MBSA = Microscopy Bulk Simple Annotations; PM = Parametric Map; ROI = region of interest; SEG = Segmentation; SR = Structured Report; TIL = tumor-infiltrating lymphocyte.

## Capabilities enabled by the harmonized DICOM representation

The harmonized representation of image-derived data in DICOM enables a number of new capabilities, which we describe below. To illustrate them, we provide a publicly available Google Data Studio[57] dashboard for browsing and filtering image-derived data across collections, and a set of Google Colaboratory notebooks demonstrating how to access, parse, and analyze the data programmatically using Python (see Code Availability).

### Unambiguous identification and formal linkage

In their original formats, image-derived data and individual spatial regions lacked persistent, globally unique identifiers. Where identifiers existed, they relied on dataset-specific conventions such as sequential row numbers or filenames. Linkage to source images was informal or implicit, relying on filenames, directory structure, or free-text references to slide names in a CSV column. The spatial relationship between image-derived data and their source images was similarly implicit. For the coordinates of spatial regions, neither the origin, nor the resolution level of the slide at which they were created, nor their units (pixel or physical) were retained within the files, but instead documented separately.

DICOM addresses these issues within the image-derived data objects themselves. First, all DICOM objects, including image-derived ones, contain unique identifiers (UIDs) at every level of the Patient-Study-Series-Instance hierarchy. Identification via UIDs also extends to finer granularities, such as individual segments in SEGs or annotation groups in MBSAs. As a consequence, whole objects as well as specific entities within them, for instance, a tumor region or a group of cells, can be unambiguously identified and referred to. This is a prerequisite for reproducible cohort building. Second, relationships between objects are encoded formally through dedicated attributes, by which derived data reference their source images and other related data. Such references can extend beyond the DICOM ecosystem to external resources, for example via digital object identifiers (DOIs), which by IDC convention we included in all generated objects. Because these linkages are machine-readable, all data derived from or otherwise related to a given slide can be identified programmatically. Last, although DICOM objects may use different resolutions and local coordinate systems, the `FrameOfReferenceUID` attribute indicates whether different images and image-derived data are defined in the same real-world coordinate system, while spatial metadata in each of them define the mapping into it. This removes ambiguities that might arise in other formats from custom coordinate conventions and provides the foundation for tools like the Slim viewer to visualize and compare spatial regions from different DICOM objects on the source or co-registered images. On the same basis, distances and areas of spatial regions can be measured in physical units, independent of the resolution level at which the regions were created.

### Semantic standardization and structured querying

Prior to harmonization, semantic information describing the meaning of image-derived data was largely absent from the data files. For example, labels describing the meaning of spatial regions were stored as free-text strings in a column of the CSV files or further information, such as units of measurements, were documented in a separate, accompanying file. Naturally, there was no standardization of these plain text labels across datasets. In raster-based formats, this problem becomes even more apparent, since a certain pixel value could represent necrosis in one dataset and stroma in another, with no information encoded in the file itself to distinguish the two.

DICOM defines an extensive set of coded concepts to describe the meaning of data, whenever possible drawing from established external terminologies such as SNOMED CT or NCIt. In the harmonized data, semantic information is therefore encoded within each image-derived data object using codes, which we applied consistently across collections. Codes are also used to describe the meaning of individual items within these objects, such as segments in SEGs or annotation groups in MBSAs, as well as accompanying measurements and their units. This enables cross-collection data operations that were not possible in the source formats. Because the codes originate from shared controlled vocabularies, image-derived data become accessible to automated structured querying and filtering at scale, regardless of the source collection or creating tool. For example, all segmentations of cell nuclei across collections can be retrieved through a single structured query, enabling cohort building based on data content rather than dataset-specific label conventions. Figure 3 shows a part of the interactive dashboard built on this capability.

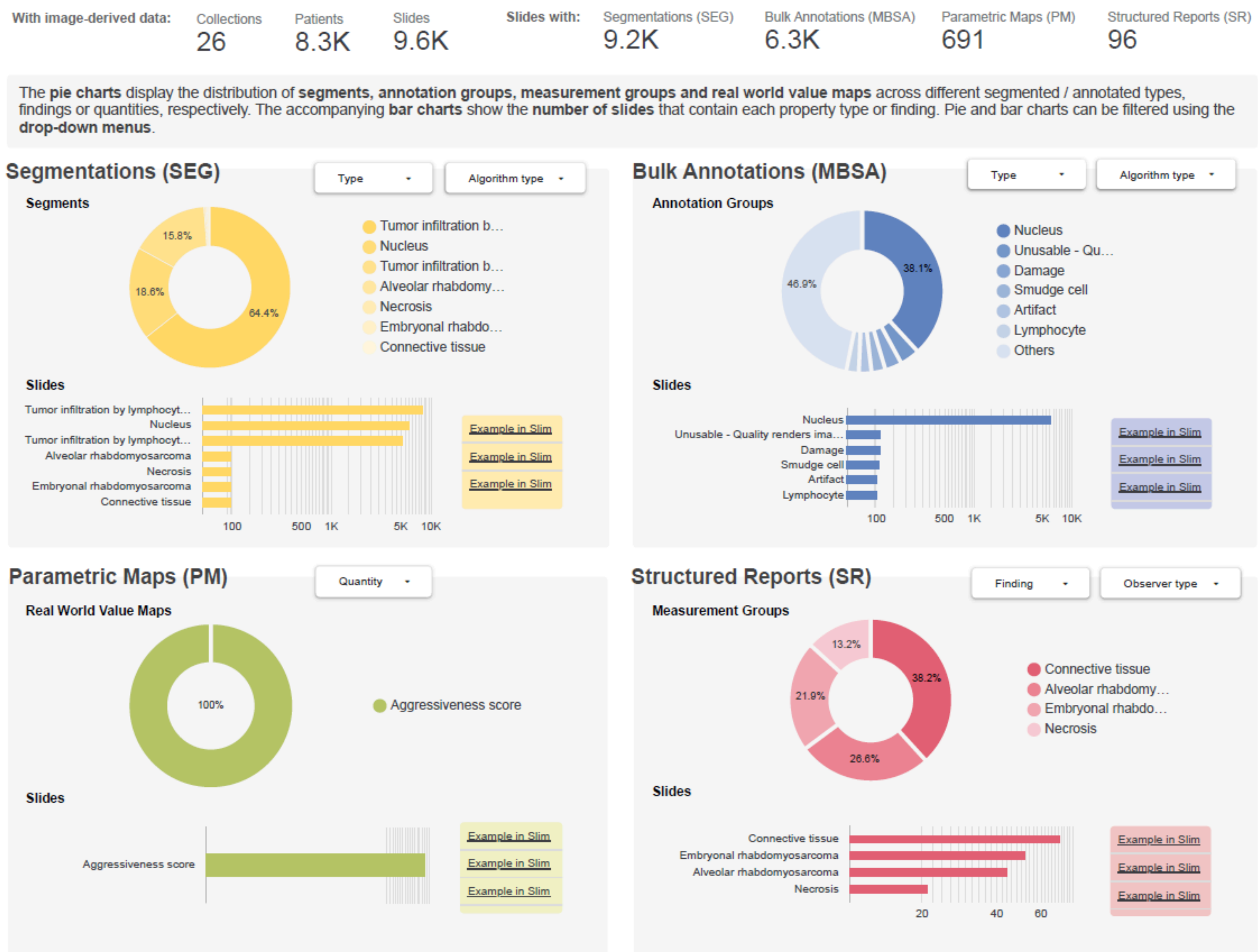

**Figure 3: Google Data Studio dashboard for visual exploration of pathology image-derived data across collections in the IDC.** The dashboard features three pages of multiple interactive charts and tables. It allows

the user to identify and subset data based on coded DICOM attributes and to open individual images and the accompanying image-derived data in the Slim viewer. This figure shows the third dashboard page. The pie charts display the distribution of segments, annotation groups, real world value maps, and measurement groups across different segmented or annotated property types, findings, or quantities. The accompanying bar charts show the number of slides that contain each property type or finding. The full interactive dashboard is publicly available at
https://datastudio.google.com/reporting/8c977edb-30b4-41df-8e53-c9662c15cb41.

### Context and provenance tracking

In their original form, the datasets recorded contextual information either in separate accompanying files or not at all, in which case it had to be retrieved from associated publications. This context included de-identified clinical information about the patient (e.g., age, diagnosis) and the study (e.g., study date), details about the slide preparation (e.g., embedding or staining), and provenance information on the image-derived data, such as whether they were created manually or algorithmically, by which institution, using what software, and when.

DICOM organizes objects within a Patient-Study-Series-Instance hierarchy. Through harmonization into DICOM, image-derived objects naturally inherit this context. In addition to this inherited context, we explicitly encoded provenance information within each object. In combination, this allows systematic evaluation of image-derived data not only by their content but also by the conditions under which they were produced and the context of the cases they belong to. For example, users can determine whether image-derived data were produced manually or algorithmically (see Figure 3, dashboard page 3), and assess whether the underlying slides match the preparation characteristics required for a particular purpose (see dashboard page 2). The accessible context is not limited to pathology. Since radiology images and their image-derived data are integrated into the same DICOM framework, cross-modality queries are possible as well.

### Uniform programmatic access

A uniform data representation across collections reduces the need for dataset-specific code. This makes it easier to apply analysis pipelines across collections, eventually leading to more reusable software and faster experimentation.

Figure 4 gives two examples using `highdicom`. Once a suitable DICOM file has been located and a coded description of the structure of interest has been chosen (for example through structured querying as described above), accessing the underlying data can proceed in a standardized way regardless of the collection and structure of interest. The first example (Figure 4a) demonstrates this for MBSA files using the SNOMED CT code for nuclei in the Pan-Cancer-Nuclei-Seg collection and for promonocytes in the BoneMarrowWSI-PediatricLeukemia collection. In both cases, graphic data are returned as a list of NumPy arrays containing the contour vertices.

In the second example (Figure 4b), we demonstrate how to access pixel data for certain segments within a fractional SEG file and format the result as a total pixel matrix. Again, segments are specified using the SNOMED CT terminology, specifically TIL presence in the TCGA-SBU-TIL-Maps collection and ARMS in the RMS-Mutation-Prediction collection.

We have prepared several Google Colaboratory notebooks that expand on the code snippets with more comprehensive examples[58]. Beyond reading and parsing, they cover downstream use cases such as estimation of cellularity in tissue areas from MBSAs (`microscopy_dicom_ann_intro.ipynb`), and the preparation of binary or label masks for machine learning model training from SRs (`RMS-Mutation-Prediction-Expert-Annotations_exploration.ipynb`) and SEGs (`microscopy_dicom_seg_intro.ipynb`). An additional notebook provides guidance on navigating the BoneMarrowWSI-PediatricLeukemia collection (`bonemarrowwsi_pediatricleukemia.ipynb`), explaining how to search for and retrieve the ROIs as well as the labeled and unlabeled hematological structures. While every notebook uses a specific collection as a concrete example, the demonstrated patterns are directly transferable to any image-derived data collection in the IDC.

a

```
import highdicom as hd

# Pan Cancer Nuclei Seg
nucleus_code = hd.sr.CodedConcept("84640000", "SCT", "Nucleus")
ann = hd.ann.annread("example.dcm")
grp = ann.get_annotation_groups(annotated_property_type=nucleus_code)[0]
contours = grp.get_graphic_data("2D")

# Bone Marrow WSI - Pediatric Leukemia
promonocyte_code = hd.sr.CodedConcept("1075005", "SCT", "Promonocyte")
ann = hd.ann.annread("example.dcm")
grp = ann.get_annotation_groups(annotated_property_type=promonocyte_code)[0]
contours = grp.get_graphic_data("2D")
```

b

```
import highdicom as hd

# TCGA SBU TIL Maps
tils_present_code = hd.sr.CodedConcept(
  "399721002", "SCT", "Tumor infiltration by lymphocytes present"
)
seg = hd.seg.segread("example.dcm")
segment_number = seg.get_segment_numbers(segmented_property_type=tils_present_code)[0]
mask = seg.get_total_pixel_matrix(segment_numbers=[segment_number])

# RMS Mutation Prediction
alveolar_rms_code = hd.sr.CodedConcept(
  "63449009", "SCT", "Alveolar rhabdomyosarcoma"
)
seg = hd.seg.segread("example.dcm")
segment_number = seg.get_segment_numbers(segmented_property_type=alveolar_rms_code)[0]
mask = seg.get_total_pixel_matrix(segment_numbers=[segment_number])
```

**Figure 4: Code to access a) graphic data from DICOM MBSA and b) pixel masks from DICOM SEG objects using `highdicom`**. Abbreviations: MBSA = Microscopy Bulk Simple Annotations; SEG = Segmentation.

Since DICOM is not merely a file format but also a communication standard, using it to encode image-derived data ensures compatibility with existing commercial and open-source DICOM-based archive and network communication software[59]. This enables standard access over a network and adds capabilities such as metadata-based querying and frame-level access[21]. Standardized network access is also the foundation for viewing data across collections in the Slim web viewer[55]. Slim uses a standard DICOMweb implementation to query available data for a specific study, identify a source image and its associated image-derived objects, and retrieve spatial metadata from each to correctly align overlays.

## Enhancements to open-source tooling

Beyond the specific collections released in the IDC, another major contribution of this work is the enhancement of the open-source software ecosystem enabling the use of image-derived DICOM objects in pathology. At the outset of this initiative, no existing library for creating and parsing such objects supported features critical for use in pathology, such as tiling and multi-resolution pyramids. Existing open-source viewers also had significant limitations. We focused our development efforts on two permissively licensed open-source software projects: the `highdicom` library and the Slim web viewer. The resulting improvements make it considerably easier for researchers to work with the collections we have released and to perform similar conversions in the future.

The `highdicom` Python library provides an API for creating and parsing image-derived DICOM objects in a way that is fully interoperable with both Python's numerical processing ecosystem (through `numpy`) and DICOM ecosystem (through `pydicom`). The following major enhancements were made to `highdicom` in the course of the work described in this paper (spanning `highdicom` versions 0.21.0 to 0.28.0):

- **Efficiency:** Due to the often enormous size of WSIs, computational pathology applications are highly sensitive to the computational efficiency of implementations. To make it practical to process the large numbers of slides in this project, we substantially improved the efficiency of `highdicom`'s implementation of creating and parsing the Segmentation IOD, reducing both number of operations and memory usage.
- **Tiling:** Computational pathology relies on tiled image representations. To simplify working with these, we implemented automatic tiling of a total pixel matrix upon construction of SEGs and PMs, as well as the reverse operation (assembling the total pixel matrix from its tiles) upon reading. Both tiled-full and tiled-sparse representations are fully supported. Users can thus work with the more natural total pixel matrix while the underlying tiled representation remains largely hidden as an implementation detail.
- **Multi-resolution pyramids:** We added support for creation of a multi-resolution pyramid series of SEGs or PMs given the highest-resolution total pixel matrix.
- **Labelmap segmentations:** We added a new implementation of SEGs with `SegmentationType LABELMAP` (see Technical considerations below for further discussion).
- **MBSA parsing:** In order to simplify working with MBSAs in analysis pipelines, we added methods to the library's MBSA implementation to find and filter annotation groups and access arrays of graphic data and measurements as `numpy` arrays.
- **CIELab color space support:** DICOM uses the CIELab color space to encode display colors for vector ROIs. To simplify its use, we added utilities to convert to and from RGB colors.
- **Documentation:** We considerably improved the library's user guide, including details of all features listed above.

The Slim web viewer is an open-source, permissively licensed DICOM viewer for pathology WSIs and associated image-derived data, implemented in Javascript. It is used as the primary viewer for pathology images on the IDC web portal. In the course of our work, we made the following improvements to Slim:

1. **De-coupling resolutions of image and image-derived data**: Slim originally required SEGs and PMs to have resolution levels that exactly matched the source image pyramid, that is, identical pixel spacings and origins at each level. This works well for segmentations that are generated at full resolution and then downsampled to the source pyramid levels. However, many derived images of clinical or research interest are created at lower resolution, such as patch-level classifications, attention maps, or saliency maps. For these, creating pyramid layers that match those of the source image would mean upsampling a small derived image into a multi-gigabyte file, which is impractical and would falsely imply spatial precision that the original measurement does not possess. Instead, such data should be stored at their native resolution, even if a matching level is absent from the source pyramid. Slim was therefore modified to support SEGs and PMs whose resolution levels do not match any level of the source pyramid.

2. **MBSA clustering**: DICOM MBSAs can contain hundreds of thousands of ROIs, ranging from individual point detections through bounding boxes to complex polygonal geometries. Rendering all of these as full vector graphics at low magnification causes severe performance degradation in the browser. To address this problem, Slim was modified to aggregate ROIs using spatial clustering. When the number of nearby ROIs exceeds a certain threshold, they are consolidated into a single cluster. When zooming in, the viewer switches to rendering them individually.

## Technical considerations

The harmonization of image-derived data into DICOM involves a range of non-trivial design decisions, each with potential implications for aspects such as downstream usability, storage efficiency or computational performance. To our knowledge, this work represents the first attempt to distribute pathology image-derived information in DICOM format. The following paragraphs summarize our key insights and recommendations for those attempting similar work in the future. Where possible, concrete guidance is provided, otherwise challenges and trade-offs are discussed. Figure 5 illustrates the main steps of our decision process when encoding image-derived data.

**Raster vs vector representation:** A fundamental decision when encoding image-derived data in DICOM is whether to represent the spatial information in vector or raster form. DICOM supports vector forms with SRs and MBSAs and raster forms with SEGs and PMs. The choice depends largely on the form in which the data were collected and on the intended (or likely) downstream uses, which commonly include interactive viewing, modification by an expert, computation of derived measures such as shape features or area fractions, and use as training targets for computational models. For each of these two aspects, one representation is usually the more practical fit, but they do not always coincide.

Vector representations are predominantly generated by manual or semi-manual annotation tools. They are a natural form for iterative manual editing, since vertices can be directly manipulated in these tools. Because they are organized by object rather than by pixel, each annotated instance (e.g., a single cell) remains individually addressable, which benefits any downstream analysis organized around discrete objects. Being resolution independent, vector representations can be projected easily and without information loss to other levels of a multi-resolution pyramid, as coordinates simply need to be multiplied by a scaling factor. This benefits both viewers and object detection models.

By contrast, raster masks are commonly generated and consumed by computational tools, including convolutional and vision transformer neural networks. Their native unit of organization is the pixel, so all annotations within a given region of the corresponding slide can be read out directly, which makes them well-suited as training targets for patch-based models. Raster masks are, however, harder to edit manually, and they are resolution dependent. Transferring them to another level of the multi-resolution pyramid, as for example required in an interactive viewer, involves resampling and interpolation, which introduces information loss.

The initial choice matters because conversion between vector and raster forms is neither entirely straightforward nor perfectly reversible. For high-resolution WSIs, such conversions can also be very computationally demanding and memory-intensive. Another consideration is file size. The size of uncompressed raster masks scales with their resolution and spatial extent, whereas the size of vector data scales with the number of contours and their level of detail (and therefore number of vertices). Very coarse regions with few vertices, such as the manual annotations in the RMS-Mutation-Prediction-Expert-Annotations collection, can be stored in small vector format files, but for more detailed contours, such as those in the Pan-Cancer-Nuclei-Seg-DICOM collection, files can become larger than the corresponding rasterized representation. Raster masks can additionally benefit from image-specific lossless compression, whereas vector representations are limited to general-purpose compression schemes such as the dictionary-based deflate algorithm[28].

For the datasets described in this paper, we encoded data in raster or vector format according to how they were submitted by the creators, in line with the IDC's role as an archive of data. Where we expected it to add further utility for downstream analysis, we additionally provided the complementary representation (e.g., Pan-Cancer-Nuclei-Seg-DICOM).

**Multi-resolution pyramids for segmentations:** For rendering of high-resolution raster masks in a viewer, we found that the use of multi-resolution pyramids was essential to achieve reasonable performance. For example, the nuclei segmentation masks in the Pan-Cancer-Nuclei-Seg-DICOM collection span tens of thousands of pixels per dimension. Loading and rendering the entire mask at full resolution leads to high network latency, exhausts browser memory, and blocks interaction. Multi-resolution pyramids allow the viewer to request only the tiles corresponding to the visible viewport at the appropriate zoom level, so that when zoomed out, the viewer displays a downsampled representation and loads full-resolution data only for the region under inspection. Viewers benefit most when the image pyramid of SEGs and PMs mirrors that of the underlying slide, with matching pixel spacing and tile dimensions at each level, yet neither the DICOM standard nor Slim (as discussed above) require this.

**Coordinate systems:** In both SRs and MBSAs, DICOM allows two different representations of coordinates[32]: image-relative coordinates, defined in pixel units relative to the top left corner of a particular slide image (in a specific plane and at a specific resolution), and coordinates relative to a frame of reference, defined in millimetres relative to an origin in physical space[60]. We chose to use image-relative coordinates, since this is how annotation tools typically supply coordinates, hence there is no mathematical imprecision in the conversion, and it is conceptually a simpler representation. This preference is also expressed by implementers in the DICOM WSI Connectathons, a series of interoperability testing events for digital pathology[59]. However, a significant advantage of coordinates relative to a frame of reference is that they are valid across multiple images defined within that same frame of reference, most pertinently different levels of a multi-resolution image pyramid. They therefore represent a good choice in situations where downstream analysis is likely to be using multiple pyramid levels, registered images (e.g. different stains of the same section), or derived images, such as segmentation masks.

**Tiled-full vs tiled-sparse:** Consistent with the experience of early adopters of DICOM for WSI[21,24,61], we confirmed that the tile-level metadata required by the tiled-sparse representation can become impractically large for SEGs and exceed metadata size limitations in some DICOM archive software, leading to errors. We therefore opted for tiled-full in all cases and recommend it as a generally more practical representation unless segmentations are extremely sparse.

**Binary vs labelmap segmentations:** At the beginning of this work, only a single form for discrete DICOM segmentations was defined: the binary segmentation type, which stores each segment as a separate one-bit-per-pixel mask and therefore allows overlapping segments. However, we found that the requirement to represent overlapping segments is relatively uncommon in practice. Meanwhile, the poor storage efficiency of binary segmentations, especially with many segments, makes them slow and unwieldy to work with. The bit-packed encoding of the pixel data, which combines multiple single-bit pixels into one byte and is unique to binary segmentations, is also more complex to implement and complicates efficient frame-level access (despite extensions to and clarifications of the DICOM standard[62–64], which may not be implemented in archives). These limitations motivated us to propose an alternative representation that is commonly used in the visualization, computer vision and image annotation communities, the labelmap[65]. This representation combines all segments into a single array in which a pixel's value encodes the index of the single segment to which that pixel belongs. As a result of our work, the labelmap segmentation type was formally adopted by the DICOM standard via supplement 243 in the 2024c edition of DICOM[27]. It was therefore not yet available when the relevant datasets were converted.

We expect that the labelmap representation will store multi-segment segmentations with far greater space efficiency than the binary form. As an additional side effect, (lossless) compression of the 8- or 16-bit pixels used in labelmap is commonly implemented across DICOM libraries and toolkits, whereas compression of the single-bit pixels used by the binary representation is defined in DICOM but to our knowledge is not currently implemented in any toolkits (work is underway to implement this at the time of writing). This leads to a significant further reduction in file size when comparing labelmap to binary forms. We found situations in which a losslessly compressed labelmap segmentation is between 14 and 48 times smaller than the equivalent uncompressed binary segmentation. We therefore plan to use the labelmap representation for some of the future pathology datasets.

**Multiple vector graphics representations:** With SRs and MBSAs, DICOM defines two IODs which can store image-derived pathology data in vector form. The fundamental difference between the two is the way metadata about each ROI are stored. SRs include metadata for each region individually. They are therefore well suited for relatively small numbers of regions (up to a few tens) that may be heterogeneous (e.g., representing different tissue types) and may be accompanied by various metadata and qualitative or quantitative evaluations. However, with high numbers of ROIs, SRs become large and slow to parse. By contrast, MBSAs factor out duplicated metadata and store the ROIs in compact arrays of appropriate precision, handling groups of homogeneous ROIs far more efficiently. In this way, MBSAs can store hundreds of thousands of ROIs and are therefore a suitable format for ROIs on the cellular or subcellular level. Measurements for each region can also be stored compactly, but only if these measurements are identical in meaning across all of them (e.g., area of the ROI). If measurements exist for only some ROIs, the `AnnotationIndexList` attribute must identify those regions, adding a layer of indirection to the authoring, validation, and consumption in viewers or analysis tools. Further qualitative or categorical information for the ROIs cannot be described within MBSAs. In practice, if SRs are not a viable alternative, encoding additional ROI metadata may therefore require workarounds – as in the BoneMarrowWSI-PediatricLeukemia collection, where we stored identifiers in the `MeasurementsSequence`.

In summary, SR should be preferred when the number of ROIs is manageable, and the evaluations are qualitative or the measurements are heterogeneous or complex to describe. MBSA should be preferred when there are many ROIs, with or without measurements, and no need to identify or describe the regions individually. For a small number of simple, homogeneous ROIs, either object is suitable (see Figure 5).

It should also be noted that SRs are extremely flexible. This flexibility can be constrained through the use of templates and selected coded concept value sets; however, most implementations support only a subset of all possible templates and their features. TID 1500 in particular is recommended by the IHE Radiology AI Results Integration Profile[31], though there are no specific recommendations for pathology yet. Interoperability between two implementations, for example the producer and consumer of an SR, therefore depends on their supported subsets being mutually compatible. This complicates interoperability in practice compared to MBSAs (as well as SEGs and PMs), which are more constrained, leave less to implementers' choice, and are therefore less likely to create compatibility issues.

**Inclusion of offset tables:** WSIs, SEGs, and PMs may all store their pixel data in compressed form. For this case, DICOM defines two (mutually exclusive) offset tables, the Basic Offset Table and the Extended Offset Table, that may be included in the file to record the location of each frame's compressed pixel data within its byte stream. Neither is required by the DICOM standard, but we strongly recommend including one, as this significantly speeds up random access to individual frames. This benefits file-based viewers and computational analysis pipelines, especially in files with many thousands of frames. Offset tables are not needed for efficient random access to frames over a network using DICOMweb, since the server is implicitly required to maintain its own index to implement the frame-level access API.

**Usage of controlled terminologies:** Throughout project execution and manual annotation of slides for the BoneMarrowWSI-PediatricLeukemia dataset, German terminology was used, necessitating the retrospective translation and mapping of 49 highly specific labels to suitable coded concepts. This proved non-trivial and required close collaboration with the hematological experts who had assigned the labels in the first place. Future similar efforts should therefore, whenever possible, employ codes from standard terminologies directly during data acquisition. As many terminologies are multilingual, annotators can work in their own language while the international code is stored, resolving translation issues from the start. Nevertheless, this approach may still face limitations in research contexts where the required terms are not yet established in clinical practice or incorporated into a standard terminology. For these cases, we recommend private codes[32] as a last resort.

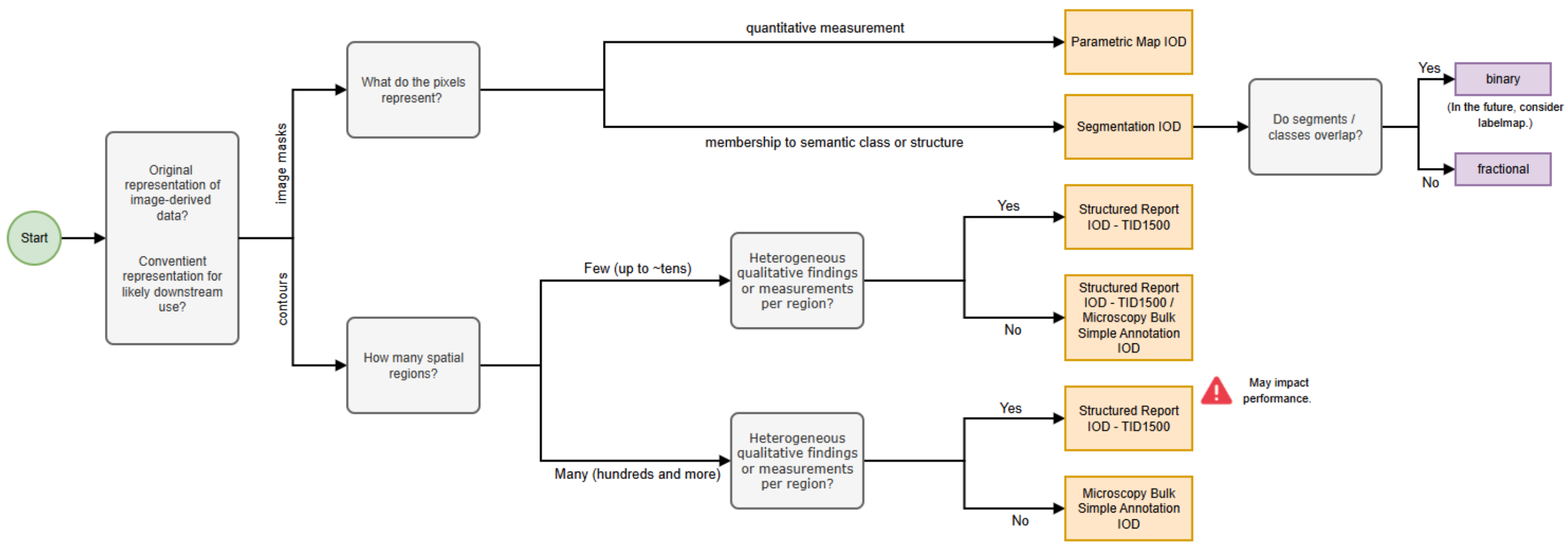

**Figure 5: Flowchart used to select the most suitable IOD.** Each decision node (rounded grey rectangle) contains a question, and the outgoing branches represent possible answers. Starting on the left (green node), the questions and answers lead to the final recommendation of the most suitable IOD (yellow rectangles) or variant of an IOD (purple rectangles).

## Limitations

During this work, we identified the following limitations both of the DICOM standard objects and in its current implementations. The former can be resolved through Supplements or Correction Proposals (CPs) to the DICOM standard, the latter through tooling improvements.

- Unlike some alternative formats like GeoJSON, DICOM MBSAs and SRs currently lack the ability to encode more complex geometry types, for example regions that contain holes. Work is in progress to extend the MBSA IOD to allow such regions via DICOM Supplement 255[66].
- The Parametric Map IOD lacks standardized attributes to store information on the algorithms or software used to generate them. DICOM CP 2652 proposes adding an Algorithm Identification Sequence to PMs to address this limitation[67].
- DICOMweb does not yet define an API for spatial querying of MBSAs or SRs from DICOM servers, for example, for retrieving only those graphic data that intersect a specified region of the underlying slide. Viewers and computational analysis tools must therefore retrieve the entire, potentially very large, set of graphic data even if only a small part is needed. This can significantly increase latency and memory requirements.
- Related to the previous point, there is no convention for how graphic data in MBSAs and SRs should be represented at reduced levels of detail, nor a DICOMweb API to retrieve such a representation. Consequently, at low magnification each viewer must apply its own simplification or subsampling heuristics, so the same data may be rendered differently across tools.
- Historically, the large size of binary SEGs has been an important limitation and disadvantage relative to alternative raster formats for segmentations. Broadening adoption of labelmap SEGs (see Technical considerations) should alleviate this problem, though this will require improved support across tools and applications.
- While semantic segmentation assigns each pixel to a class without distinguishing individual objects, instance segmentation additionally preserves each object's identity, which is relevant, for example, for counting mitotic figures[68]. However, representing instance segmentations using SEGs is impractical, since the size of the descriptive metadata grows unwieldy with large numbers of separately identified instances, regardless of whether these are encoded as binary masks or labelmaps.

# Discussion

To date, image-derived pathology data are still shared only rarely, and without any shared convention. We have harmonized five such datasets from different source formats into standard DICOM representation and have made them available via the IDC. The converted collections constitute a resource in their own right, as they enrich 26 existing image collections with a range of scientifically valuable image-derived information that can now be searched, accessed, and visualized alongside them. To our knowledge, this is the first public release of large-scale scientific data derived from pathology images in DICOM format. Through this work, we have demonstrated that the existing DICOM standard and our implementations have the capability to support a range of real-world image-derived data, covering ROIs and dense quantifications with varied semantics and granularity, generated by both human experts and artificial intelligence tools. Furthermore, vector and raster representations for ROIs are both supported, allowing selection of the representation best suited for a particular use case, for example vector data for visualization or raster data for training deep learning models. Together, this establishes the feasibility of our approach for future data sharing programs. At the same time, harmonizing the source data into DICOM required a considerable effort to collect metadata describing the content, which was often missing from the original representation, and entailed a series of design decisions and technical trade-offs as detailed in the Results section. Addressing these challenges has led to essential improvements to reusable open-source tooling (`highdicom` and Slim) that will hopefully benefit similar efforts and catalyze further development of the DICOM pathology ecosystem.

As demonstrated in the Results section, the use of DICOM for encoding image-derived pathology data has several important advantages compared to the source representations. These are (1) unambiguous identification, linkage and spatial alignment between derived data and their source images; (2) semantic standardization, enabling structured querying of the data within and across collections; (3) integration of contextual and provenance information; and (4) uniform programmatic access which simplifies the software ecosystem. Beyond these, DICOM is natively integrated into the clinical care environment, in contrast to alternative standard formats such as those of the OME ecosystem[69–72]. Aligning formats between clinical and research settings holds the potential to accelerate the translation of research tools into clinical practice, simplify collaborations between physician-scientists and computational researchers, and facilitate the use of de-identified clinically-produced image data for development and testing. Moreover, since radiology images and their derived data already operate within the DICOM framework, adopting the same standard for pathology allows all data to be indexed, stored, and presented alongside each other[73], substantially reducing the technical overhead of cross-modality studies.

We recognize that using DICOM in place of general-purpose formats, such as CSV, PNG, or GeoJSON, is not without challenges. Loading and interpreting the content of DICOM files requires specialized libraries. While numerous established open-source libraries for parsing DICOM attributes exist (e.g., `DCMTK` in C++, `pydicom` in Python, and `dcmjs`[74] in JavaScript), these offer a rather low-level interface that can be difficult to use for imaging researchers without expertise in DICOM. Open-source libraries with higher-level APIs that simplify parsing and writing of complex image-derived objects, such as the ones we used, are only now emerging. Among these, `highdicom` currently offers the most extensive support. At the same time, relatively few tools support visualization of image-derived DICOM objects. In radiology, established viewers such as OHIF[75] and 3D Slicer[47,76] can read and write DICOM SEG and SR TID1500 objects, whereas equivalent support in pathology is only emerging now. Adoption of MBSA can be observed through the DICOM WSI Connectathons, where WSI scanner vendors and software developers jointly test the interoperability of their implementations[59]. Slim addresses the remaining gap by providing a reference viewer implementation that covers all object types used in this work.

This work opens up several directions for future development. First, the datasets presented in this work do not exercise all capabilities of DICOM for encoding image-derived data. Slide-level information, for instance biomarker scores or quality metrics, can already be represented in DICOM using SR, and demonstrating their utility in the future could be a valuable next step. Second, further work is necessary to fully realize all benefits of DICOM for pathology image-derived data in practice. In particular, broader support in up- and downstream libraries to create, read, or display the respective DICOM objects remains critical for wider adoption, for example, in popular open-source tools like QuPath[77], OpenSlide[78], or `wsidicom`[79]. Here, `highdicom` and Slim can serve as reference implementations, and the collections released as test data for validating new ones. We also plan to make our conversion code more generic and reusable, lowering the barrier for others to convert their existing image-derived datasets into DICOM in the future. Third, the DICOM standard itself is expected to evolve through the Standard's established change-proposal and supplement process to address the limitations we encountered (see Limitations). Ultimately, the translation of DICOM pathology image-derived data into broader research and clinical practice will require coordinated testing efforts across the community. As mentioned above, DICOM WSI Connectathons have already extended their scope to include image-derived content in the form of MBSAs[59]. Sustaining this effort and extending it to SRs, SEGs, and PMs would be an important next step toward broad, robust interoperability in the field.

# Conclusion

In this manuscript, we have presented our approach to harmonizing five datasets of research image-derived pathology data into DICOM and have made them publicly available within the National Cancer Institute (NCI) Imaging Data Commons (IDC). The converted collections span a wide variety of image-derived content, covering semantic classes (structures, tissue regions, tumor types), scales (coarse slide regions, image patches, and subcellular structures), data representations (raster, vector), and sources (generated manually and computationally), demonstrating the flexibility of the DICOM format and providing a reference set of examples for further tool development. By co-locating image-derived data alongside the original images in the IDC, in a single shared form, we have enriched those images and streamlined reuse for further research projects. Little consensus currently exists about how to store and communicate pathology image-derived data for clinical or research use. Our efforts have addressed this gap by demonstrating that DICOM, in addition to standing out for its compatibility with existing clinical systems, is a viable format for this purpose and warrants consideration when planning how analysis results are stored and shared. Through our work, and that of others, open-source viewers and libraries are available for working with and visualizing image-derived data in computational pathology, and these serve as an important foundation on which further tools can be built for scalable and interoperable management across diverse imaging datasets.

# Data Availability

The converted image-derived data collections are accessible from the IDC portal at the following URLs. Each of them is also accompanied by a Zenodo record.

- Pan-Cancer-Nuclei-Seg-DICOM[80]:
    - https://zenodo.org/records/14009675
    - https://portal.imaging.datacommons.cancer.gov/explore/filters/?analysis_results_id=Pan-Cancer-Nuclei-Seg-DICOM
- RMS-Mutation-Prediction-Expert-Annotations[81]:
    - https://zenodo.org/records/14941043
    - https://portal.imaging.datacommons.cancer.gov/explore/filters/?analysis_results_id=RMS-Mutation-Prediction-Expert-Annotations
- TCGA-SBU-TIL-Maps[82]:
    - https://zenodo.org/records/16966286
    - https://portal.imaging.datacommons.cancer.gov/explore/filters/?analysis_results_id=TCGA-SBU-TIL-Maps
- TCGA-GBM360[83]:
    - https://zenodo.org/records/17470191
    - https://portal.imaging.datacommons.cancer.gov/explore/filters/?analysis_results_id=TCGA-GBM360
- BoneMarrowWSI-PediatricLeukemia[52]:
    - https://zenodo.org/records/18499180
    - https://portal.imaging.datacommons.cancer.gov/explore/filters/?collection_id=Community&collection_id=bonemarrowwsi_pediatricleukemia

The Google Data Studio dashboard for exploration of the converted image-derived data collections can be accessed here: https://datastudio.google.com/reporting/8c977edb-30b4-41df-8e53-c9662c15cb41

# Code Availability

The code for the dataset conversions is publicly available at the following Github repositories:

- BoneMarrowWSI-PediatricLeukemia collection (includes code for conversion of source images in MIRAX format and the conversion of image-derived data from CSV): https://github.com/ImagingDataCommons/conversion_mirax_dicom.

- All remaining image-derived data collections: https://github.com/ImagingDataCommons/idc-sm-annotations-conversion.

Both repositories are under active development; the specific versions used to create the data described in the paper are archived on Zenodo[37,38].

The code for `highdicom` and the Slim viewer is also publicly available at Github:

- `highdicom`: https://github.com/ImagingDataCommons/highdicom
- Slim: https://github.com/ImagingDataCommons/slim

The Google Colaboratory notebooks illustrating the use of DICOM SEGs, SRs, and MBSAs from the IDC are available in the IDC-Tutorials Github repository[58]:

- Introduction to DICOM MBSAs: https://github.com/ImagingDataCommons/IDC-Tutorials/blob/master/notebooks/pathomics/microscopy_dicom_ann_intro.ipynb
- Introduction to DICOM SRs for computational pathology: https://github.com/ImagingDataCommons/IDC-Tutorials/blob/master/notebooks/collections_demos/rms_mutation_prediction/RMS-Mutation-Prediction-Expert-Annotations_exploration.ipynb
- Introduction to DICOM SEGs for computational pathology: https://github.com/ImagingDataCommons/IDC-Tutorials/blob/master/notebooks/pathomics/microscopy_dicom_seg_intro.ipynb
- Introduction to the BoneMarrowWSI-PediatricLeukemia collection: https://github.com/ImagingDataCommons/IDC-Tutorials/blob/master/notebooks/collections_demos/bonemarrowwsi_pediatricleukemia.ipynb

# Funding

This project has been funded in whole or in part with Federal funds from the National Cancer Institute, National Institutes of Health, under Task Order No. HHSN26110071 under Contract No. HHSN261201500003I.

# Supplementary Information

## Supplement 1: Links to images used in Figure 2 of the main article

The images in Figure 2 of the main text were generated using the IDC's public instance of the Slim web viewer. The specific cases used in the figure may be reached at the following public URLs.

Pan-Cancer-Nuclei-Seg-DICOM:
https://viewer.imaging.datacommons.cancer.gov/slim/studies/2.25.2340910091818719684056189239962087328 70/series/1.3.6.1.4.1.5962.99.1.1661307024.1148056373.1638043879568.2.0

RMS-Mutation-Prediction-Expert-Annotations:
https://viewer.imaging.datacommons.cancer.gov/slim/studies/2.25.1550174847564987304921365972389948388 76/series/1.3.6.1.4.1.5962.99.1.3434988325.342625608.1687062201125.4.0

TCGA-SBU-TIL-Maps:
https://viewer.imaging.datacommons.cancer.gov/slim/studies/2.25.3100736045752592779566040365787961613 38/series/1.3.6.1.4.1.5962.99.1.1977422950.1768973582.1638359995494.2.0

TCGA-GBM360:
https://viewer.imaging.datacommons.cancer.gov/slim/studies/2.25.6880309589696627658338213892496483927 4/series/1.3.6.1.4.1.5962.99.1.1163866303.1057408148.1637546438847.2.0

BoneMarrowWSI-PediatricLeukemia:
https://viewer.imaging.datacommons.cancer.gov/slim/studies/1.2.826.0.1.3680043.8.498.4936208324755901841272041118405493518/series/1.2.826.0.1.3680043.8.498.85786393993207301181346242160854391271

## Table S1: Codes for the labeled hematological structures in the BoneMarrowWSI-PediatricLeukemia collection.

| | **English free-text label in the source dataset** | **Code for `AnnotationPropertyCategory`** | **Code for `AnnotationPropertyType`** |
|---|---|---|---|
| 1 | artifact | SCT:260787004 \|Physical object\| | SCT:47973001 \|Artifact\| |
| 2 | basophilic_band | SCT:91723000 \|Anatomical structure | LN:LP65588-3 \| Basophils.band form |
| 3 | basophilic_erythroblast | SCT:91723000 \|Anatomical structure | SCT:464005 \|Basophilic erythroblast (cell)\| |
| 4 | basophilic_metamyelocyte | SCT:91723000 \|Anatomical structure | SCT:63369000 \| Basophilic metamyelocyte (cell)\| |

| 5 | basophilic_myelocyte | SCT:91723000 \|Anatomical structure | SCT:17295002 \| Basophilic myelocyte (cell)\| |
|---|---|---|---|
| 6 | damaged_cell | SCT:49755003 \|Morphologically abnormal structure (morphologic abnormality)\| | SCT:37782003 \|Damage\| |
| 7 | degranulated_neutrophilic_metamyelocyte | SCT:91723000 \|Anatomical structure | NCIt:C37174 \| Neutrophil with Cytoplasmic Hypogranularity |
| 8 | degranulated_neutrophilic_myelocyte | SCT:91723000 \|Anatomical structure | SCT:250292003 \|Hypogranular white blood cell\| |
| 9 | eosinophilic_band | SCT:91723000 \|Anatomical structure | LN:LP65365-6 \| Eosinophils.band form |
| 10 | eosinophilic_metamyelocyte | SCT:91723000 \|Anatomical structure | SCT:80036001 \|Eosinophilic metamyelocyte (cell)\| |
| 11 | eosinophilic_myelocyte | SCT:91723000 \|Anatomical structure | SCT:90961003 \|Eosinophilic myelocyte (cell)\| |
| 12 | erythrocyte | SCT:91723000 \|Anatomical structure | SCT:41898006 \|Erythrocyte (cell)\| |
| 13 | giant_platelet | SCT:49755003 \|Morphologically abnormal structure (morphologic abnormality)\| | SCT:44687006 \|Giant platelet (morphologic abnormality)\| |
| 14 | lymphoid_precursor_cell | SCT:91723000 \|Anatomical structure | SCT:127910003 \|Lymphoid precursor cell\| |
| 15 | lymphocyte | SCT:91723000 \|Anatomical structure | SCT:56972008 \|Lymphocyte (cell)\| |
| 16 | lymphocytic_blast | SCT:91723000 \|Anatomical structure | SCT:15433008 \|Lymphoblast (cell)\| |
| 17 | lymphoidocyte | SCT:91723000 \|Anatomical structure | NCIt:C12847 \| Reactive Lymphocyte |
| 18 | macrophage | SCT:91723000 \|Anatomical structure | SCT:58986001 \|Macrophage (cell)\| |
| 19 | megakaryocyte | SCT:91723000 \|Anatomical structure | SCT:23592000 \|Megakaryocyte (cell)\| |
| 20 | micromegakaryocyte | SCT:91723000 \|Anatomical structure | SCT:33196003 \|Micromegakaryocyte (cell)\| |
| 21 | mitosis | SCT:91723000 \|Anatomical structure | SCT:75167008 \|Mitotic cell (cell)\| |

| 22 | immature_monocyte | SCT:91723000 \|Anatomical structure | NCIt:C13120 \| Immature Monocyte |
|---|---|---|---|
| 23 | monocyte | SCT:91723000 \|Anatomical structure | SCT:55918008 \|Monocyte (cell)\| |
| 24 | monocytic_blast | SCT:91723000 \|Anatomical structure | SCT:53945006 \|Monoblast (cell)\| |
| 25 | myeloid_precursor_cell | SCT:91723000 \|Anatomical structure | SCT:127914007 \|Myeloid precursor cell\| |
| 26 | myelocytic_blast | SCT:91723000 \|Anatomical structure | SCT:15622002 \|Myeloblast (cell)\| |
| 27 | neutrophil_extracellular_trap | SCT:91723000 \|Anatomical structure | NCIt:C180931 \| Neutrophil Extracellular Trap |
| 28 | neutrophilic_band | SCT:91723000 \|Anatomical structure | SCT:702697008 \|Band neutrophil (cell)\| |
| 29 | neutrophilic_metamyelocyte | SCT:91723000 \|Anatomical structure | SCT:50134008 \|Neutrophilic metamyelocyte (cell)\| |
| 30 | neutrophilic_myelocyte | SCT:91723000 \|Anatomical structure | SCT:4717004 \|Neutrophilic myelocyte (cell)\| |
| 31 | orthochromatic_erythroblast | SCT:91723000 \|Anatomical structure | SCT:113334004 \|Orthochromic erythroblast (cell)\| |
| 32 | phagocytosis | SCT:91723000 \|Anatomical structure | SCT:56639005 \|Phagocytosis, function (observable entity)\| |
| 33 | plasma_cell | SCT:91723000 \|Anatomical structure | SCT:113335003 \|Plasma cell (cell)\| |
| 34 | polychromatic_erythroblast | SCT:91723000 \|Anatomical structure | SCT:16779009 \|Polychromatophilic erythroblast (cell)\| |
| 35 | proerythroblast | SCT:91723000 \|Anatomical structure | SCT:16671004 \|Proerythroblast(cell)\| |
| 36 | prolymphocyte | SCT:91723000 \|Anatomical structure | SCT:19394005 \|Prolymphocyte (cell)\| |
| 37 | promegakaryocyte | SCT:91723000 \|Anatomical structure | SCT:50284009 \|Promegakaryocyte (cell)\| |
| 38 | promonocyte | SCT:91723000 \|Anatomical structure | SCT:1075005 \|Promonocyte (cell)\| |
| 39 | promyelocyte | SCT:91723000 \|Anatomical structure | SCT:43446009 \|Promyelocyte (cell)\| |
| 40 | pseudo_gaucher_cell | SCT:91723000 \|Anatomical | SCT:59870003 \|Gaucher-like |
| | | structure | cell (cell)\| |
| 41 | segmented_basophil | SCT:91723000 \|Anatomical structure | SCT:30061004 \|Basophil, segmented (cell)\| |
| 42 | segmented_eosinophil | SCT:91723000 \|Anatomical structure | SCT:14793004 \|Eosinophil, segmented (cell)\| |
| 43 | segmented_neutrophil | SCT:91723000 \|Anatomical structure | SCT:80153006 \|Segmented neutrophil (cell)\| |
| 44 | smudge_cell | SCT:49755003 \|Morphologically abnormal structure (morphologic abnormality)\| | SCT:34717007 \|Smudge cell (morphologic abnormality)\| |
| 45 | spicule | SCT:91723000 \|Anatomical structure | NCIt:C82998 \| Spicule |
| 46 | technically_unfit | SCT:309825002 \|Spatial and relational concepts\| | DCM:111235 \| Unusable - Quality renders image unusable |
| 47 | thrombocyte | SCT:91723000 \|Anatomical structure | SCT:16378004 \|Thrombocyte |
| 48 | thrombocyte_aggregate | SCT:91723000 \|Anatomical structure | SCT:60649002 \|Platelet aggregation\| |
| 49 | unknown_blast | SCT:91723000 \|Anatomical structure | SCT:312256009 \|Blast cell\| |
| 50 | no_consensus_found | SCT:91723000 \|Anatomical structure | SCT:414387006 \|Structure of hematological system\| |

**Table S1: Coded values for the labeled hematological structures in the BoneMarrowWSI-PediatricLeukemia collection.** The original labels were translated from German to English and mapped to suitable coded values with support from clinical experts involved in the project. The English translation of the original label was used to encode the free-text DICOM attribute `AnnotationGroupLabel`.